\documentclass[5p,twocolumn]{elsarticle}

\usepackage{hyperref}
\usepackage{amsmath}
\usepackage{graphicx}
\usepackage{subfigure}
\usepackage{multirow}
\usepackage{color}
\journal{Journal of \LaTeX\ Templates}

\hypersetup{draft}
\begin{document}

\begin{frontmatter}

\title{Attend To Count: Crowd Counting with Adaptive Capacity Multi-scale CNNs}
% \tnotetext[mytitlenote]{ Corresponding author}

\author[huake]{Zhikang Zou}
\address[huake]{School of Electronic Information and Communication, Huazhong University of Science and Technology}

\author[dalao]{Yu Cheng}
\address[dalao]{Microsoft Research \& AI}

\author[huake]{Xiaoye Qu}

\author[zheda]{Shouling Ji}
\address[zheda]{College of Computer Science and Technology, Zhejiang University}

\author[ibm]{Xiaoxiao Guo}
\address[ibm]{IBM \& AI Foundations Learning}

\author[huake]{Pan Zhou\corref{cor1}}
\cortext[cor1]{Corresponding author}

% \author[mymainaddress,mysecondaryaddress]{Elsevier Inc}

% \author[mysecondaryaddress]{Global Customer Service\corref{mycorrespondingauthor}}
% \cortext[mycorrespondingauthor]{Corresponding author}

% \address[mymainaddress]{1600 John F Kennedy Boulevard, Philadelphia}
% \address[mysecondaryaddress]{360 Park Avenue South, New York}

\begin{abstract}
Crowd counting is a challenging task due to the large variations in
crowd distributions. Previous methods tend to tackle the whole image
with a single fixed structure, which is unable to handle diverse
complicated scenes with different crowd densities. Hence, we propose
the Adaptive Capacity Multi-scale convolutional neural networks
(ACM-CNN), a novel crowd counting approach which can assign
different capacities to different portions of the input. The
intuition is that the model should focus on important regions of the
input image and optimize its capacity allocation conditioning on the
crowd intensive degree. ACM-CNN consists of three types of modules:
a coarse network, a fine network, and a smooth network. The coarse
network is used to explore the areas that need to be focused via
count attention mechanism, and generate a rough feature map. Then
the fine network processes the areas of interest into a fine feature
map. To alleviate the sense of division caused by fusion, the smooth
network is designed to combine two feature maps organically to
produce high-quality density maps. Extensive experiments are
conducted on five mainstream datasets. The results demonstrate the
effectiveness of the proposed model for both density estimation and
crowd counting tasks.
\end{abstract}


\end{frontmatter}

%\linenumbers

\section{Introduction}

\begin{figure}
\centering
\includegraphics[width=8.5cm]{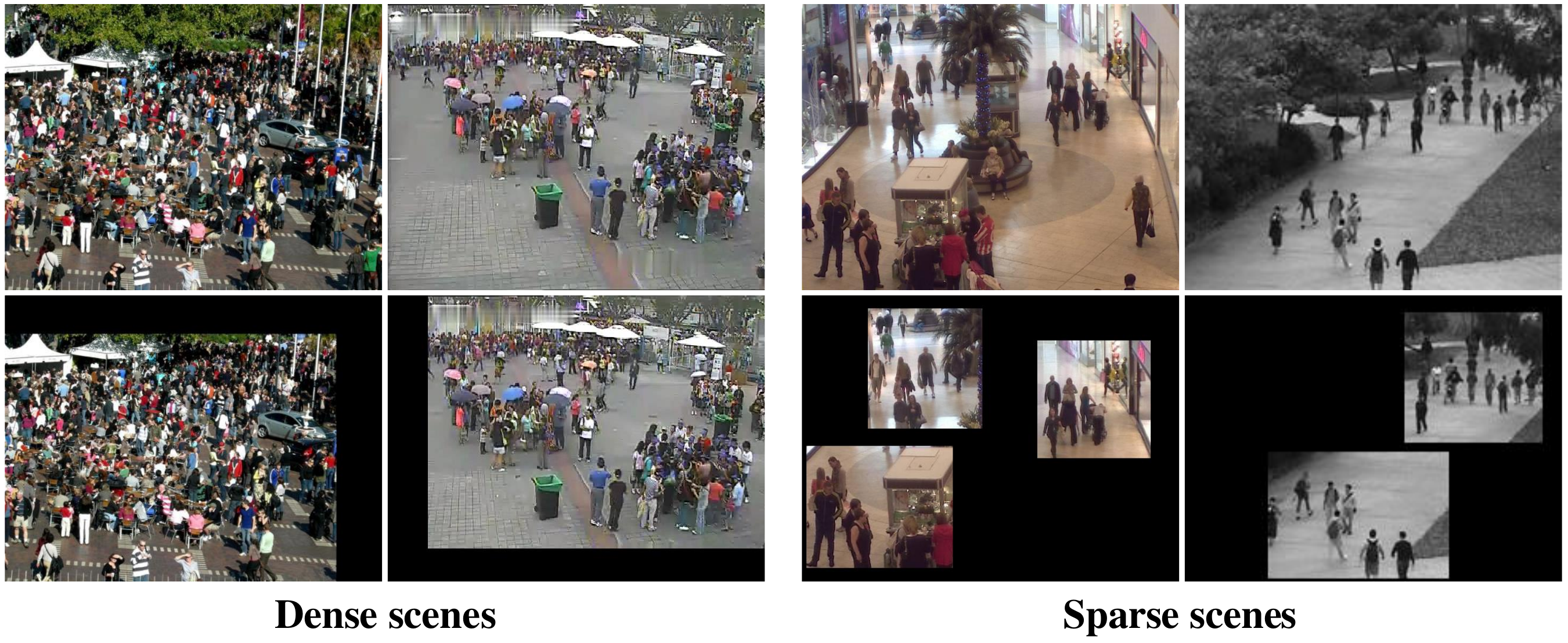}
\caption{Top row indicates typical static crowd scenes from the most
commonly used datasets (Shanghaitech, WorldExpo'10, Mall, UCSD in
order) and the bottom row represents their corresponding attention
map from our network.} \label{example}
\end{figure}

The goal of crowd counting is to count the number of crowds in a
surveillance scene, which lays on an important component in many
computer vision applications. As the first and the most important
part of crowd management, automatic crowd counting can monitor the
crowd density of surveillance areas and alert the manager for safety
control if the density exceeds specified thresholds. However,
precise crowd estimation remains challenging as it demands the
extractor to be able to capture pedestrians in various scenes with
diverse population distribution (see Figure \ref{example}).

% Early methods are generally classified into detection and regression based methods. The detection based methods localize pedestrian's position with object detectors, but it is extremely limited by scene density because the counting performance will decrease as the scene become more crowded. On the contrary, the regression based methods directly learning the mapping {\color{red}{from}} the image features to the total counts. However, extracting hand-craft features is not efficient for consuming many resources.

Previous methods \cite{viola2003detecting,KOCAK2017105} adopt a
detection-style framework, where a sliding window detector is used
to estimate the number of people. The limitation of such
detection-based methods is that severe occlusion among people in a
clustered environment always results in poor performance. To deal
with images of dense crowds, some focus on regression-based methods
\cite{chan2009bayesian,chen2012feature}, which directly learn a
mapping from the features of image patches to the count in the
region. Despite the progress in addressing the issues of occlusion,
these methods are not efficient since extracting hand-craft features
consumes many resources.

In the past decades, the approaches built on Convolutional Neural
Networks (CNN) have shown strong generalization ability to handle
complicated scenarios by virtue of automatic feature extraction
process. However, the performance of different models varies largely
across different crowd densities. MCNN \cite{zhang2016single},
utilizing three simple columns with different receptive fields to
tackle the scale variations, performs well in sparsely populated
scenes such as UCSD dataset but loses its superiority in dense crowd
scenes like Shanghai dataset. Whereas SaCNN
\cite{DBLP:journals/corr/abs-1711-04433} that incorporates a deep
backbone with small fixed kernels turns out to be the opposite.
% This indicates that a single model is not compatible with all datasets due to multiplex density levels.
This indicates that a single model is not sufficient enough to
effectively cope with all datasets with multiple density levels.
Even for images in the same dataset, the distribution of crowds is
not uniform, which means processing the whole image with one fixed
model usually
% results in over-estimating or under-predicting.
results in under-estimating or over-estimating count in the crowd
image. The work Switch-CNN \cite{sam2017switching} divides the crowd
scenes into non-overlapping patches and sends each patch to a
particular column depending on the classification network. Although
this strategy improves the adaptability of the model to some extent,
% the irrationality of artificially divided areas, the limited enumeration of several receptive fields and the similar ability of the same capacity in each column hinders the improvement.
there are two main drawbacks in this work. One is that the
representation ability is similar for each column with the same
capacity, the other is that artificially dividing images is not
proper.

One obvious solution to address these problems is to assign adaptive
capacities to different portions across the scene
% since certain regions represent a particular type of crowd.
based on their density levels. The aforementioned examples
illustrate that a deep network is able to handle the regions with
high density while a shallow network achieves better performance in
sparse regions. Therefore, the goal is divided into two aspects: (1)
a model is built to automatically distinguish areas of different
density levels in the scene, as is shown in Figure \ref{example};
(2) networks of different capacities are specialized for specific
areas and integrated to obtain the final result.

To achieve this, we propose the Adaptive Capacity Multi-scale CNNs
(ACM-CNN), which could focus on important regions via count
attention mechanism without prior knowledge and assign its capacity
depending on the density level. This is achieved by exploiting three
subnetworks: a coarse network, fine network and smooth network. The
coarse network is a multi-column architecture with shallow layers,
which is used to locate dense regions and generate a rough feature
map. While the fine network, a deep architecture based on VGGNets
\cite{DBLP:journals/corr/SimonyanZ14a}, processes the dense regions
into a fine feature map. Since adding up two feature maps directly
will lead to the sense of division in the result images, we
incorporate the smooth network to fuse two features organically and
thereby generate the high-quality density map. ACM-CNN is a fully
convolutional design which can be optimized via an end-to-end
training scheme.

In summary, the main contributions of this paper are:
\begin{itemize}
\item As far as we know, our work is the first attempt to introduce the adaptive capacity model for crowd counting problem. The primary aim is to take advantage of the different complex networks, which have unique representation abilities to deal with the regions with different density levels.
\item We propose a novel attention mechanism termed as count attention, which could automatically locate dense regions of the input without prior knowledge.
\item We propose a general crowd counting system that can choose the proper network of the right capacity for a specific scene of crowd distribution.
% \item We evaluate ACM-CNN on five large-scale benchmark datasets, Shanghaitech, UCSD, Mall, UCF\_CC\_50 and World Expo'10. The experimental results show that our model outperforms all state-of-the-art approaches. Besides, comparative experiments are designed to verify the effectiveness and versatility of our count attention.
\end{itemize}

\section{Related work}
\subsection{Crowd counting}
Crowd counting has attracted lots of researchers to create various
methods in pursuit of a more accurate result
\cite{DBLP:journals/corr/abs-1903-03303,8486651,7927432,ZHANG2019144,SINDAGI20183}.
The earlier traditional methods \cite{KOCAK2017105} which adopt a
detection-style framework have trouble with solving severe
occlusions and high clutter. To overcome this issue,
regression-based methods \cite{chan2009bayesian,chen2012feature}
have been introduced. The main idea of these methods is to learn a
mapping between features extracted from the local images to their
counts. Nevertheless, the representation ability of the low-level
features are limited, which cannot be widely applied. Recently, most
researchers focus on Convolutional Neural Network (CNN) based
approaches inspired by the great success in visual classification
and recognition
\cite{DBLP:journals/corr/abs-1903-00853,DBLP:journals/corr/abs-1811-11968,DBLP:journals/corr/abs-1904-01333}.
In order to cope with the scale variation of people in crowd
counting, the MCNN \cite{zhang2016single} use receptive fields of
different sizes in each column to capture a specific range of head
sizes. With similar idea, Sam \textit{et al.}
\cite{sam2017switching} train a switch classifier to choose the best
column for image patches while Sindagi \textit{et al.}
\cite{Sindagi_2017_ICCV} encode local and global context into the
density estimation process to boost the performance. Further, Sam
\textit{et al.} \cite{Sam_2018_CVPR} extend their previous work by
training a growing CNN which can progressively increase its
capacity. Later, scale aggregation modules are proposed by Cao
\textit{et al.} \cite{Cao_2018_ECCV} to improve the representation
ability and scale diversity of features. Instead of using
multi-column architectures, Li \textit{et al.} \cite{Li_2018_CVPR}
modify the VGG nets with dilated convolutional filters to aggregate
the multi-scale contextual information. Liu \textit{et al.}
\cite{Liu_2018_CVPR} adaptively adopt detection and regression count
estimations based on the density conditions. Differently, some works
pay more attention to context information. Ranjan \textit{et al.}
\cite{Ranjan_2018_ECCV} propose an iterative crowd counting
framework, which first produces the low-resolution density map and
then uses it to further generate the high-resolution density map.
Further, multi-scale contextual information is incorporated into an
end-to-end trainable pipeline CAN
\cite{DBLP:journals/corr/abs-1811-10452}, so the proposed network is
capable of exploiting the right context at each image location.

\begin{figure*}[!htb]
\centering
\includegraphics[height=10cm,width=16.5cm]{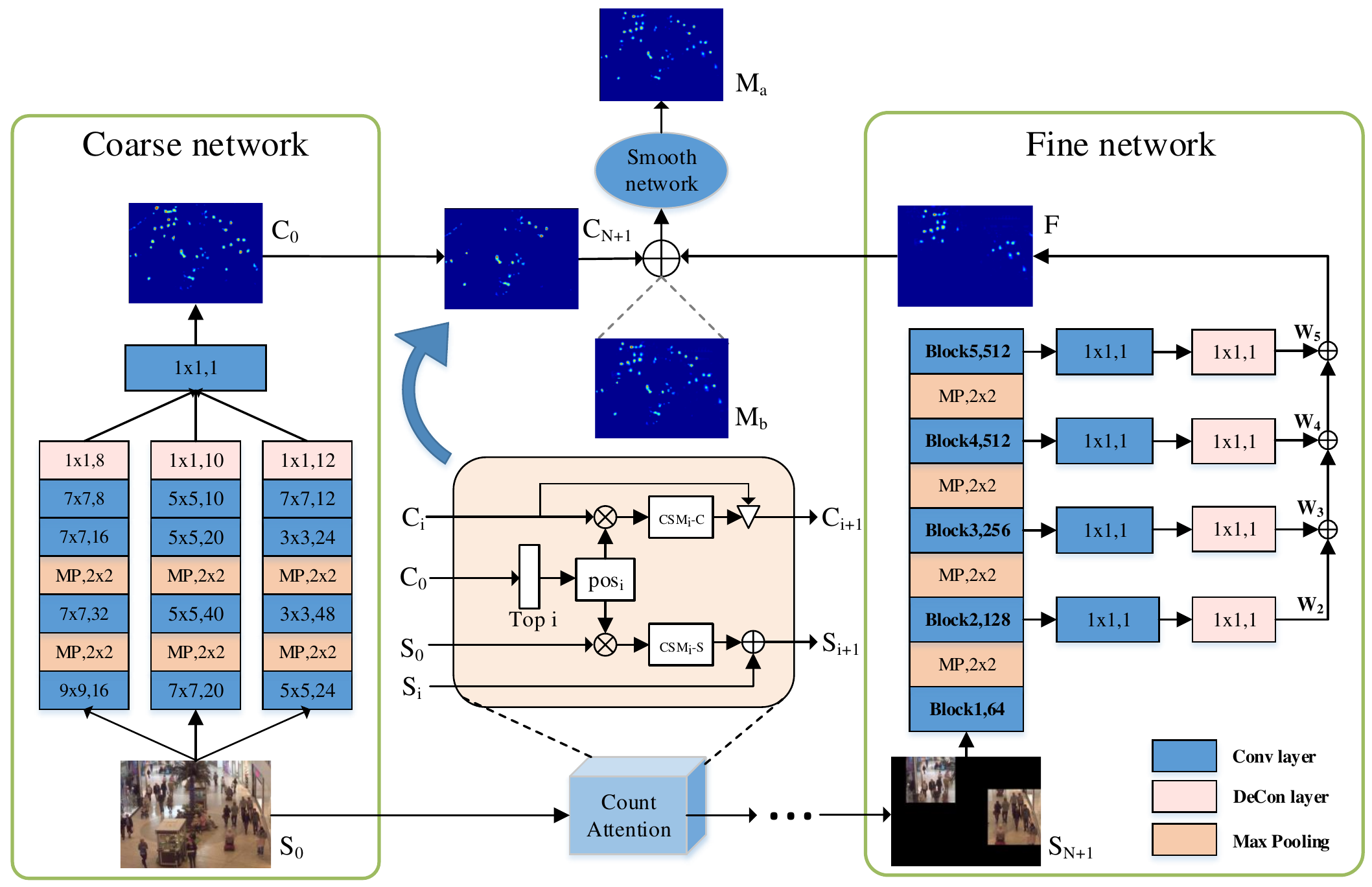}
\caption{Architecture of the proposed Adaptive Capacity Multi-scale
convolutional neural networks. The coarse network takes the input
image $S_0$ to generate initial feature map $C_0$. The count
attention module iteratively processes $C_0$ to locate specific
high-density $S_{N+1}$. The fine network takes $S_{N+1}$ to produce
detailed map $F$ then merged with $C_{N+1}$. Finally, the fusion
feature map $M_b$ is relayed to a smooth network to generate
prediction density map $M_a$.} \label{wholestructure}
\end{figure*}

\subsection{Attention mechanism} Recent years have witnessed the boom of deep convolutional nerual network in many challenging tasks, ranging from image classification to object detection \cite{chen2016deeplab,kruthiventi2017deepfix,JI2018130,GEPPERTH201614,shuangjiejointly}. However, it is computationally expensive because the amount of the computation scales increases linearly with the number of image pixels. In parallel, the concept of attention has gained popularity recently in training neural networks, allowing models to learn alignments between different modalities \cite{DBLP:journals/corr/MnihHGK14,GanCKLLG19}. In \cite{DBLP:journals/corr/BaMK14}, researchers attempt to use hard attention mechanism learning to selectively focus on task-relevant input regions, thus improving the accuracy of recognizing multiple objects and the efficiency of computation. Almahairi \textit{et al.} \cite{DBLP:journals/corr/AlmahairiBCZLC15} propose dynamic capacity network based on hard attention mechanism that efficiently identifies input regions to which the DCN's output is most sensitive and we should devote more capacity. Recently, some researchers attempt to use the attention mechanism in crowd counting. MA Hossain \textit{et al.} \cite{8659316} propose a novel scale-aware attention network by combining both global and local scale attentions. In addition, a novel dual path multi-scale fusion network architecture with attention mechanism named SFANet \cite{DBLP:journals/corr/abs-1902-01115} is proposed, which can perform accurate count estimation for highly congested crowd scenes. Inspired by these works above, we prove that it is effective to introduce the attention mechanism in crowd counting because networks with different complexity have their unique characterization capabilities for different population density regions. Therefore, we propose a new attention strategy named count attention that assigns different capacities to different portions of the input.

\section{Methodology}

\subsection{Count Attention}
As discussed in Introduction, different types of networks have their
specific learning ability in areas with different crowd density,
hence our primary task is to differentiate regions of various
density levels. In order to introduce the attention strategy, we
must first describe how to conduct the data processing, which is
converting an image with labeled people heads to a density map.

Suppose the head at pixel $x_i$, it can be represented as a delta
function $\delta(x-x_i)$. Therefore, we can formulate the image
$I(x)$ with $N$ labeled head positions as:
\begin{center}
\begin{equation}
I(x) = \sum_{i=1}^{N}{\delta(x-x_i)}
\end{equation}
\end{center}
Due to the fact that the cameras are generally not in a bird-view,
the pixels associated with different locations correspond to
different scales. Therefore, we should take the perspective
distortion into consideration. Following the method introduced in
the MCNN \cite{zhang2016single}, the geometry-adaptive kernels are
applied to generate the density maps. Specifically, the ground truth
$D(x)$ is computed by blurring each head annotation with a Gaussian
kernel $G_\sigma$ normalized to one, which is:
\begin{equation}
D(x) = \sum_{i=1}^{N}{\delta(x-x_i)*G_{\sigma_i}(x)},
\;with\;\sigma_i = \beta\overline{d^i}
\end{equation}
where $\overline{d^i}$ is denoted as the average distance between
the head $x_i$ and its $k$ nearest neighbors, $\beta$ and $k$ are
hyper-parameters. In this paper, we set $\beta$ = 0.3 and $k$ = 3.
Thus, the variance of Gaussian kernel $\sigma_i$ is in proportion to
$d_i$. When the crowd density is larger, $d_i$ will be smaller.
Accordingly, $\sigma_i$ will be smaller. For one head, the
integration of density values equals to one, so the smaller
$\sigma_i$ means larger center value. In conclusion, the highest
values in the map indicate the most densely populated area to which
the network should pay great attention. The count attention
mechanism is to traverse all the pixels in the density map to select
a set of positions with highest pixel values and crop a series of
blocks centering on these points as the attention regions (dense
crowd regions).

\subsection{Adaptive Capacity Multi-scale CNN}
 Our model is composed of a coarse network, fine network, and smooth network as shown in Figure \ref{wholestructure}. Given an input image $S_0$, the coarse network first extracts its abstract features to generate a rough density map $C_0$. To further boost the quality of the density map $C_0$, count attention is applied to iteratively generate specific high-density level regions $S_{N+1}$. In detail, through this strategy, the geometric locations of $i^{th}$ ($i \in (1,2,...,N)$) highest pixel in $C_0$ are picked as a mapping center at each time, denoted as $pos_i$. Similarly, the corresponding spatial mapping region centering on $pos_i$ is represented as $CSM_i$. We generate a blank image $S_1$ with the same size as $S_0$ and use $S_i$ ($i \in (1,2,...,N+1)$) to represent a high-density area map generated by each iteration. Apparently, when $i$ equals 1, $S_i$ indicate the initialized blank map $S_1$. In each iteration, the $CSM_i$ of input image $S_0$ ($CSM_i-S$) fills in $S_i$, which is formulated as:
 \begin{equation}
S_{i+1} = S_i+CSM_i \otimes S_0
\end{equation}
 Here $\otimes$ means cropping the region of $S_0$ according to the corresponding spatial mapping region centering on $pos_i$. Then this process keeps loop until $i$ equals $N$. The whole procedure can be represented as:
\begin{equation}
S_{N+1} = S_1+\sum_{i=1}^{N}CSM_i\otimes S_0
\end{equation}
In this way, the high-density regions $S_{N+1}$ are selected and
then relayed to a fine network to capture high-level abstraction F.
To incorporate initial map $C_0$ with F, the low-density regions of
$C_0$ are required to be preserved while the remaining area needs to
be discarded. This step can also be synchronized with the
aforementioned count attention. We use $C_i$ ($i \in (1,2,...,N+1)$)
to represent the left density map after each iteration. When $i$
equals 1, we define $C_1$ is identical with $C_0$. In each
iteration, the $CSM_i$ of feature map $C_0$ ($CSM_i-C$) is gradually
stripped out from a feature map $C_i$, which is formulated as:
\begin{equation}
C_{i+1} = C_i-CSM_i\otimes C_0
\end{equation}
Then this process keeps loop until $i$ equals $N$. The whole
procedure can also be represented as:
\begin{equation}
C_{N+1} = C_1-\sum_{i=1}^{N}CSM_i\otimes C_0
\end{equation}
After obtaining low-density feature $C_{N+1}$, it can add up with
$F$ to obtain a more accurate feature map $M_b$. However, this
fusion map still exposes a sense of stiff merging. To solve this
issue, this fusion result $M_b$ is relayed to a smooth network
containing several convolutional layers to produce the final density
map $M_a$.

The detailed design of coarse network, fine network, and smooth
network will be introduced in the following section.

\subsection{Model Configuration}
\begin{figure}
\centering
\includegraphics[width=8.5cm]{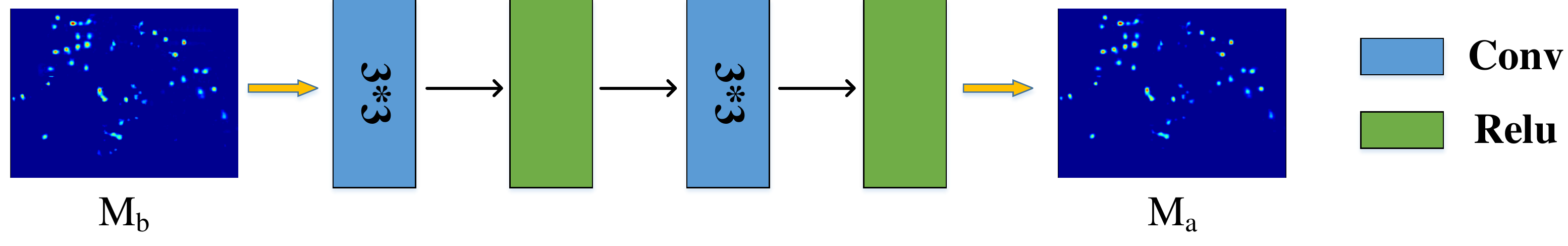}
\caption{An overview of our proposed smooth network.} \label{smooth}
\end{figure}

\noindent\textbf{Coarse network} A three-column CNN which is similar
to MCNN \cite{zhang2016single} for identical kernel size and filter
number constitutes the coarse network. However, prominent difference
resides in the deconvolution layer following each column of the
network, which up-samples to generate the feature map with the same
size as the input image, as is shown in Figure \ref{wholestructure}
pink rectangle. Compared to the down-sampling resultant map of MCNN
\cite{zhang2016single}, the deconvolution layer can ensure more
accurate positions of highest pixel values which are then picked out
as the centers of the selected high-density regions via count
attention mechanism. The coarse network is adopted for the reason of
its simplicity and effectiveness on sparse scenes. \\\textbf{Fine
network} The fine network is based on VGG-16 with all the fully
connected layers and the last pooling layer removed. Four 1*1
convolution layers followed by 1*1 deconvolution layers are
connected to the last four blocks ($\rm{Block}_{\rm{j}}$ ($j \in
(2,3,4,5)$)) respectively to deliver the different levels of
semantic information. Suppose $F_j$ ($j \in (2,3,4,5)$) represents
the output of each deconvolution layer. In order to effectively
integrate these context, four learnable weights $W_j$ ($j \in
(2,3,4,5)$) are assigned to $F_j$ ($j \in (2,3,4,5)$) respectively
for dynamically adjusting the importance of each component, which
can be formulated as:
\begin{equation}
F = \sum_{j=2}^{5}{W_{j}*F_j}
\end{equation}
Where $F$ represents the fine feature map. This strategy helps to
achieve more fine granularity crowd description. Similarly, the
choice of this fine network relies largely on its superiority
dealing with dense scenes. \\\textbf{Smooth network} The smooth
network is made up of CR(12,3) and CR(1,3), where C means
convolution layer, R refers to ReLU layer, the first number in every
brace indicates filter number, and the second number denotes filter
size. An overview of the mooth network is shown in Figure
\ref{smooth}. The use of this smooth network contributes to smoother
prediction map generation.

\subsection{Networks Optimization}
The whole model is trained on the entire dataset by back-propagating
$l_2$ loss via an end-to-end strategy. The design of our model
ensures that all the desired results have the same resolutions as
the input images. Suppose there are $N$ training images and the
parameters before the smooth network are $\Theta_m$. Then
$M_b(X_i;\Theta_m)$ represents the merged feature map before the
smooth network for the i-th input image $X_i$. We first conduct a
supervision of this intermediate result to ensure priority learning
of the coarse network and the fine network. That is:
\begin{equation}
L_b = \frac{1}{2N}\sum_{i=1}^{N}{||M_b(X_i;\Theta_m) - D(X_i)||_2^2}
\end{equation}
where $D(X_i)$ indicates the corresponding ground truth for $X_i$.
Besides, the differences between the final density map
$M_a(X_i;\Theta_m;\Theta_s)$ and the ground truth are minimized by:
\begin{equation}
L_a = \frac{1}{2N}\sum_{i=1}^{N}{||M_a(X_i;\Theta_m;\Theta_s) -
D(X_i)||_2^2}
\end{equation}
where $\Theta_s$ indicates the parameters of the smooth network. A
weighted combination is computed on the above two loss functions to
get the final objective:
\begin{equation}
L_{overall} = L_a + \lambda L_b
\end{equation}
where $\lambda$ is the hyper-parameter balancing the learning of the
first two networks and the smooth network. In our experiments, the
value $\lambda$ is set to 1. We train the proposed model using the
Adam solver with the following parameters: learning rate $10^{-5}$
and batch size $1$.

\section{Experiments}
\subsection{Evaluation Metrics}
The proposed model ACM-CNN is evaluated on four major crowd counting
datasets. Following the existing works, we adopt two standard
metrics, Mean Absolute Error (MAE) and Mean Squared Error (MSE), to
benchmark the model. For a test sequence with $N$ images, MAE and
MSE are defined as follows:
\begin{equation}
MAE = \frac{1}{N}\sum_{i=1}^{N}{\|c_i-\widetilde{c_i}\|}
\end{equation}
\begin{equation}
MSE{\rm{ = }}\sqrt {\frac{1}{N}{{\sum\limits_{i = 1}^N {\left\|
{{c_i} - \widetilde {{c_i}}} \right\|}^2 }}}
\end{equation}
where $\widetilde{c_i}$ indicates the actual count and $c_i$
represents the estimated number of pedestrians in the i-th image.
MAE reflects the accuracy of the predicted count and MSE is an
indicator of the robustness.

\begin{table}[!htb]
\newcommand{\tabincell}[2]{\begin{tabular}{@{}#1@{}}#2\end{tabular}}
% \footnotesize
%\renewcommand{\arraystretch}{1.3}
\caption{Comparisons of ACM-CNN with other state-of-the-art methods
on ShanghaiTech dataset \cite{zhang2016single}.}\label{Shanghai}
\centering
\begin{tabular}{c|c|c|c|c}
\hline
&\multicolumn{2}{c|}{Part\_A}&\multicolumn{2}{c}{Part\_B}\\
\hline
Method&MAE&MSE&MAE&MSE\\
\hline \hline
Zhang \textit{et al.} (2015)&181.8&277.7&32.0&49.8\\
\hline
MCNN (2016) &110.2&173.2&26.4&41.3\\
\hline
TDF-CNN&97.5&145.1&20.7&32.8\\
\hline
Switching-CNN&90.4&135.0&21.6&33.4\\
\hline
CP-CNN &73.6&106.4&20.1&30.1\\
\hline
SaCNN & 86.8 &  139.2 & \textbf{16.2} & 25.8 \\
\hline
ACM-CNN (ours) &\textbf{72.2}&\textbf{103.5}& 17.5 &\textbf{22.7}\\
\hline
\end{tabular}
\end{table}

\subsection{Shanghaitech dataset}
\begin{figure}
\centering
\includegraphics[width=8.8cm]{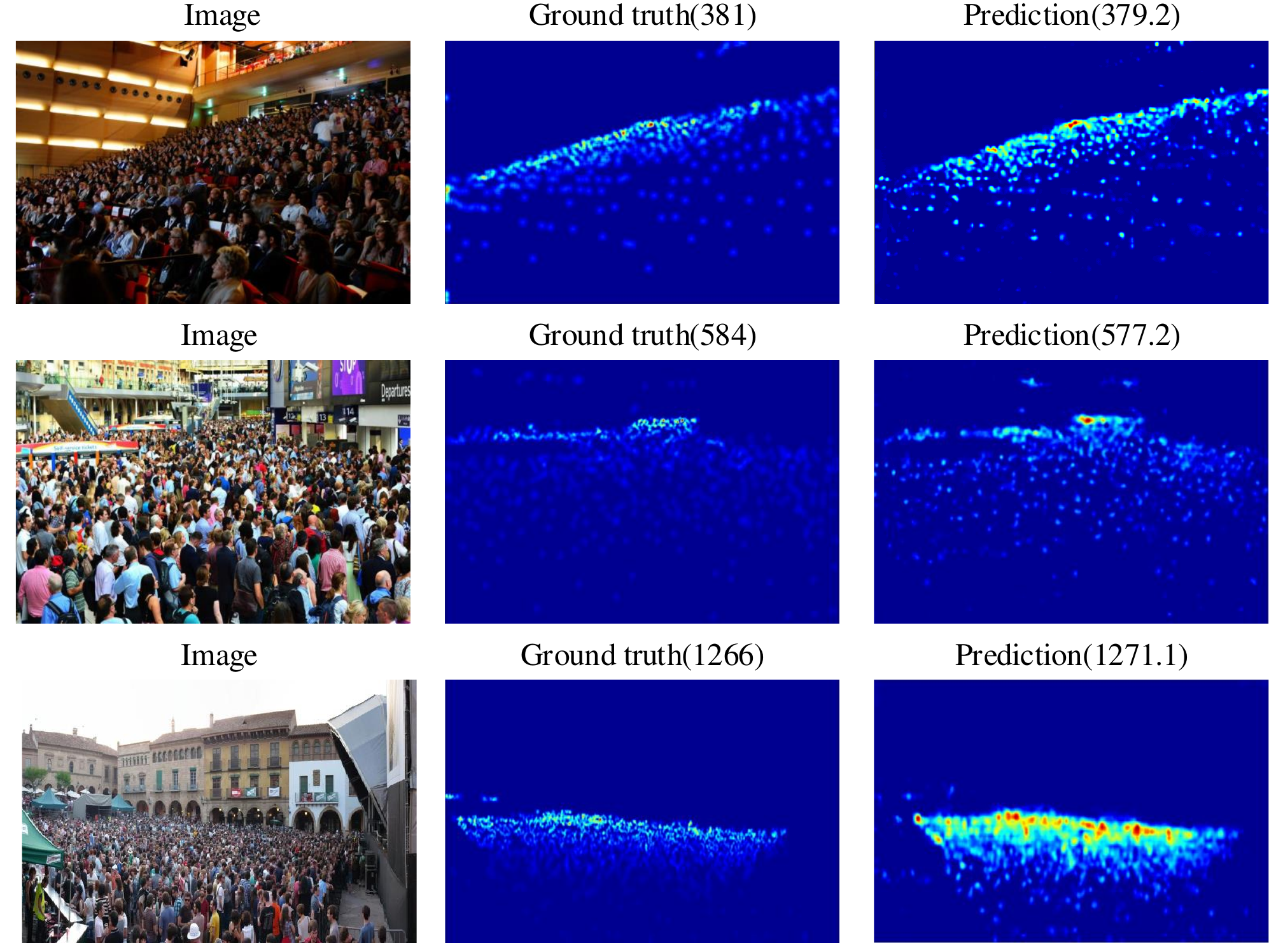}
\caption{Visualization of the crowd density maps by various
methods.} \label{shanghai_density}
\end{figure}

The Shanghaitech crowd counting dataset \cite{zhang2016single}
consists of 1198 annotated images with a total of 330,165 people,
which is said to be the largest one in terms of the number of
annotated people. It has two parts: one of the parts named Part\_A
contains 482 images which are randomly crawled from the Internet,
the other named Part\_B includes 716 images which are taken from the
busy streets of metropolitan area in Shanghai. Each of the two parts
is divided into training and testing sets: in Part\_A, 300 images
are used for training and the remaining are used for testing while
400 images of Part\_B are used for training and 316 for testing. To
augment the training set, we crop 100 patches from each image at
random locations and each patch is 1/4 size of the original image
for both Part\_A and Part\_B. We compare performance between our
approach with other state-of-the-art methods in Table
\ref{Shanghai}, including Zhang \textit{et al.}
\cite{zhang2015cross}, MCNN \cite{zhang2016single}, TDF-CNN
\cite{Sam2018TopDownFF}, Switch-CNN \cite{sam2017switching}, CP-CNN
\cite{Sindagi_2017_ICCV}, SaCNN
\cite{DBLP:journals/corr/abs-1711-04433}. The results indicate that
ACM-CNN is able to calculate the number of crowds more accurately.
We also report some samples of the test cases in Figure
\ref{shanghai_density}.

\begin{table*}[htb]
\newcommand{\tabincell}[2]{\begin{tabular}{@{}#1@{}}#2\end{tabular}}
% \footnotesize
\renewcommand{\arraystretch}{1.3}
\caption{Mean absolute errors of the WorldExpo'10 crowd counting
dataset \cite{zhang2015cross}. ACM-CNN delivers the lowest average
MAE compared to other methods.} \label{table_WorldExpo} \centering
\begin{tabular}{c|c|c|c|c|c|c}
\hline
Method & Scene1 &Scene2 & Scene3 & Scene4 & Scene5 & Average MAE\\
\hline \hline
LBP + RR \cite{zhang2016single} & 13.6 & 59.8 & 37.1 & 21.8 & 23.4 & 31.0\\
\hline
Zhang \textit{et al.}\cite{zhang2015cross} & 9.8 & 14.1 & 14.3 & 22.2 & 3.7 & 12.9\\
\hline
MCNN \cite{zhang2016single} & 3.4 & 20.6 & 12.9 & 13.0 & 8.1 & 11.6\\
\hline
Switching-CNN \cite{sam2017switching} & 4.4 & 15.7 & \textbf{10.0} & 11.0 & 5.9 & 9.4\\
\hline
CP-CNN \cite{Sindagi_2017_ICCV} & 2.9 & 14.7 & 10.5 & \textbf{10.4} & 5.8 & 8.86\\
\hline
DecideNet \cite{Liu_2018_CVPR}& \textbf{2.00} & 13.14 & 8.90 & 17.40 & 4.75 & 9.23\\
\hline \hline
ACM-CNN (ours) & 2.4 & \textbf{10.4} & 11.4 & 15.6 & \textbf{3.0} & \textbf{8.56}\\
\hline
\end{tabular}
\end{table*}

\begin{figure}
\centering
\includegraphics[width=8.5cm]{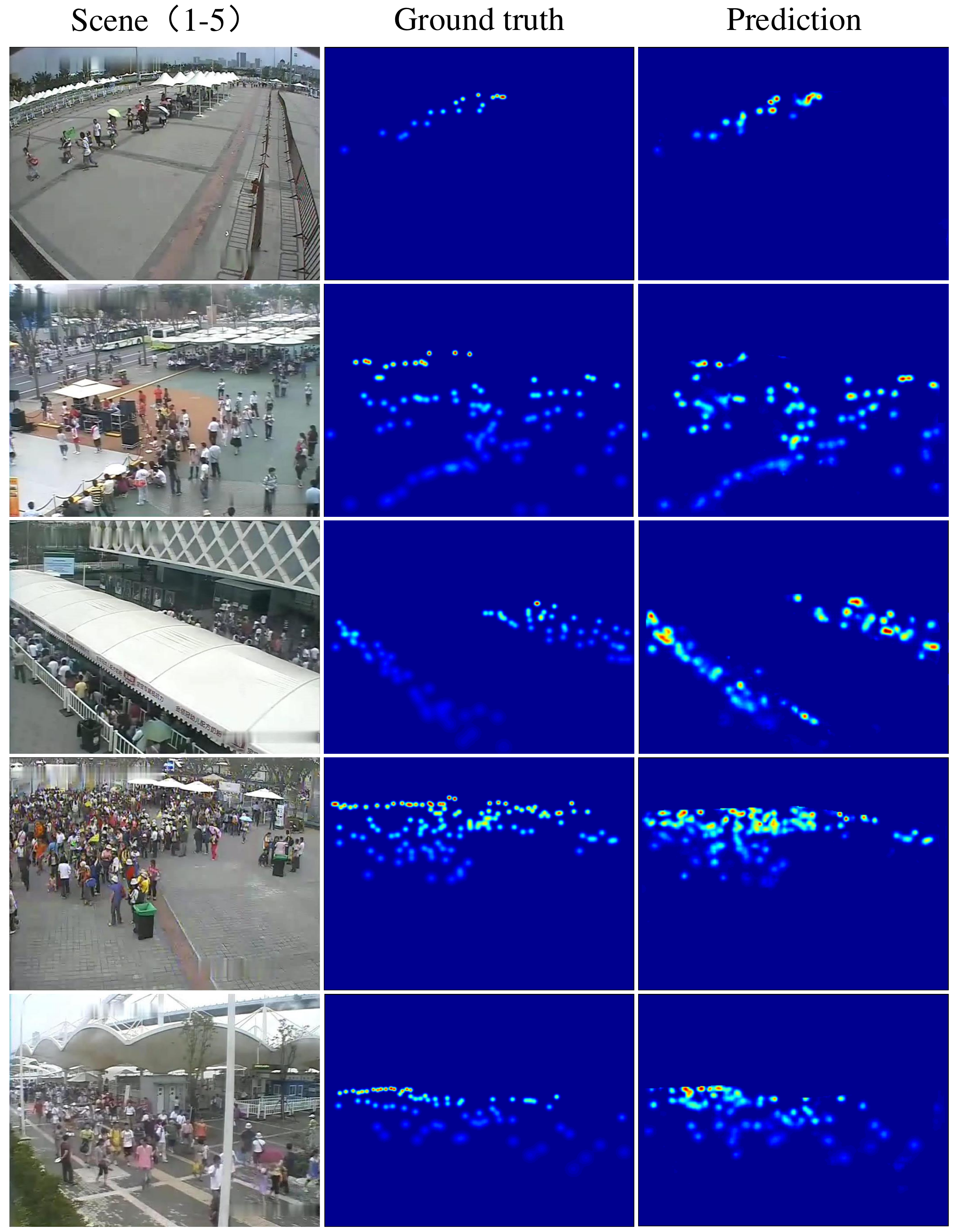}
\caption{Example results on WorldExpo'10 dataset. Each row
represents a typical scene.} \label{world_density}
\end{figure}

\begin{figure}
\centering
\includegraphics[width=8.5cm]{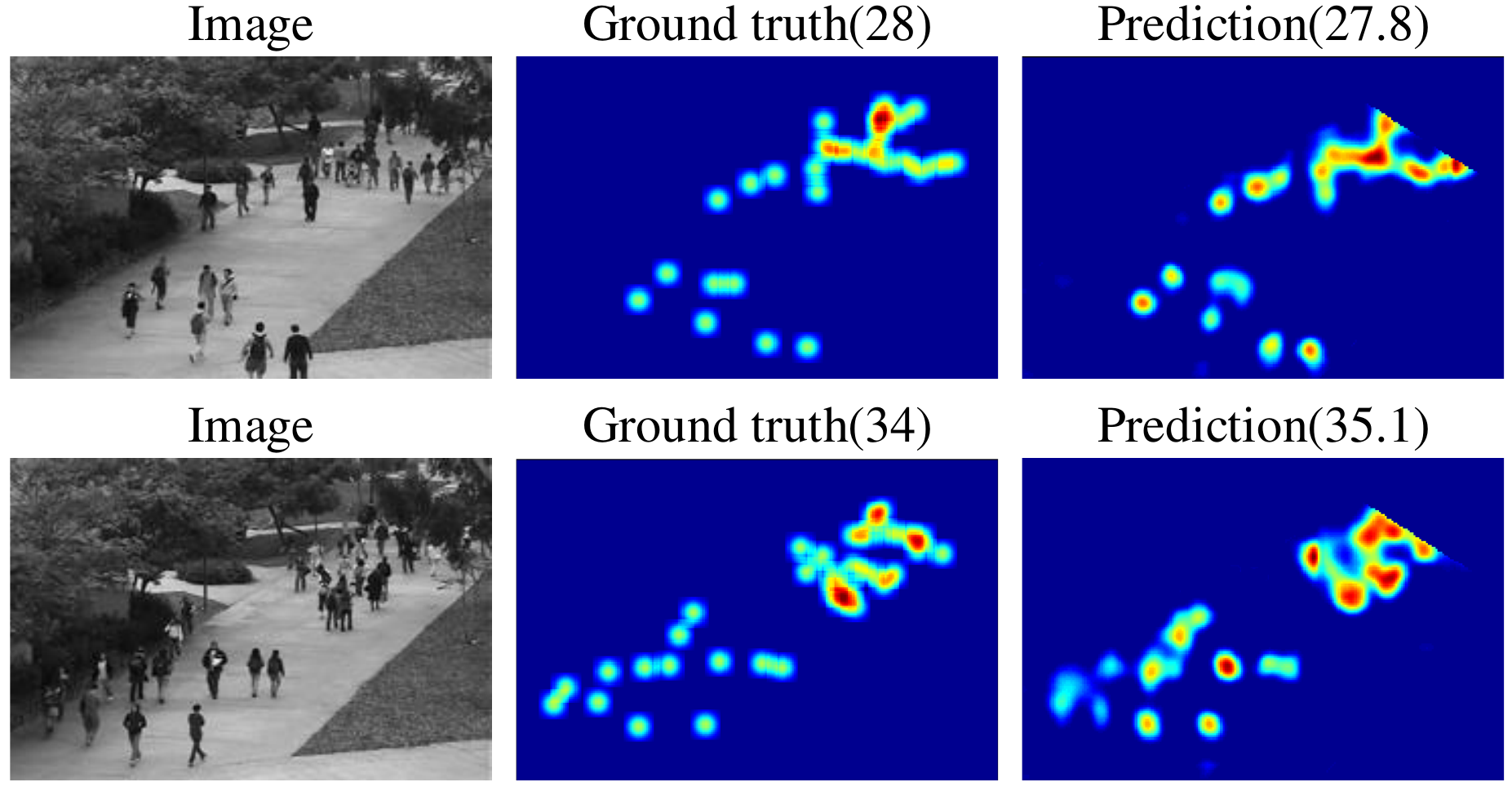}
\caption{Results on UCSD dataset using the proposed ACM-CNN.}
\label{ucsd_density}
\end{figure}

\subsection{WorldExpo'10 dataset}
The WorldExpo'10 dataset introduced by \cite{zhang2015cross}
consists of 1132 annotated video sequences that cover a large
variety of scenes captured by 108 surveillance equipment in Shanghai
2010 WorldEXPO. Each frame of one sequence with 50fps frame rate has
a regular 576 x 720 pixels. A total of 199,923 pedestrians are
labeled at the centers of their heads in 3980 frames. They also
provide ROI for each of the scenes. We split the frames into two
parts: one part with 3380 frames in 103 scenes are treated as
training and verification sets, and the other part with 600 frames
in 5 scenes are treated as test sets. In order to conduct data
augmentation, we crop 10 patches of size 256*256 per image, and the
same operation is done on the ROI (regions of interest) images. We
compare the performance of our model with other state-of-the-art
methods and the results are reported in Table \ref{table_WorldExpo}.
Also, we visualize the results of the five test scenes in Figure
\ref{world_density}.

\begin{table}[!htb]
\newcommand{\tabincell}[2]{\begin{tabular}{@{}#1@{}}#2\end{tabular}}
% \footnotesize
%\renewcommand{\arraystretch}{1.3}
\caption{Comparisons results: Estimation errors on the UCSD dataset
\cite{chan2008privacy}. Gaussian Process Regression refers to the
work in \cite{chan2008privacy}.}\label{UCSD} \centering
\begin{tabular}{c|c|c}
\hline
Method & MAE & MSE \\
\hline \hline
Gaussian Process Regression & 2.16 & 7.45 \\
\hline
Zhang \textit{et al.} (2015)&1.60& 3.31\\
\hline
MCNN (2016) &1.07&1.35 \\
\hline
Switching-CNN &1.62&2.10 \\
\hline
CSRNet & 1.16 & 1.47\\
\hline
ACM-CNN (ours) & \textbf{1.01} & \textbf{1.29} \\
\hline

\end{tabular}
\end{table}

\subsection{UCSD dataset}
Our second experiment concentrates on crowd counting for UCSD
dataset introduced in \cite{chan2008privacy}. It is acquired with a
stationary camera mounted at an elevation in UCSD campus. The crowd
density ranges from sparse to very crowded. This is a 2000-frame
video dataset that is recorded at 10 fps with a frame size of 158
*238. Different from Shanghaitech dataset above, this dataset not
only provides ground truth with figure coordinates but also region
of interest (ROI) for each frame. Of the 2000 frames, we send frames
601 through 1400 into the model for training while the remaining
frames are used for testing. There is no process of augmentation due
to the similarity of pictures. During training, we separately prune
the ground truth, merged map, and the final map with ROI. As a
result, the error is back-propagated for areas inside the ROI. Table
\ref{UCSD} lists the results of all methods and our approach is able
to offer better MAE and MSE in all scenes, which means our model can
generate a more accurate density map whether in dense or sparse
scenes. Two examples are shown in Figure \ref{ucsd_density}.

\begin{table}[!htb]
\newcommand{\tabincell}[2]{\begin{tabular}{@{}#1@{}}#2\end{tabular}}
% \footnotesize
%\renewcommand{\arraystretch}{1.3}
\caption{Performance evaluation of various methods on Mall dataset
\cite{chen2012feature}.}\label{mall} \centering
\begin{tabular}{c|c|c}
\hline
Method & MAE & MSE \\
\hline \hline
Gaussian process regression & 3.72 & 20.1 \\
\hline
Ridge regression&3.59& 19.0\\
\hline
MoCNN &2.75&13.4 \\
\hline
Count forest &2.5&10.0 \\
\hline
ACM-CNN (ours) & \textbf{2.3} & \textbf{3.1} \\
\hline

\end{tabular}
\end{table}

\begin{figure}
\centering
\includegraphics[width=8.8cm]{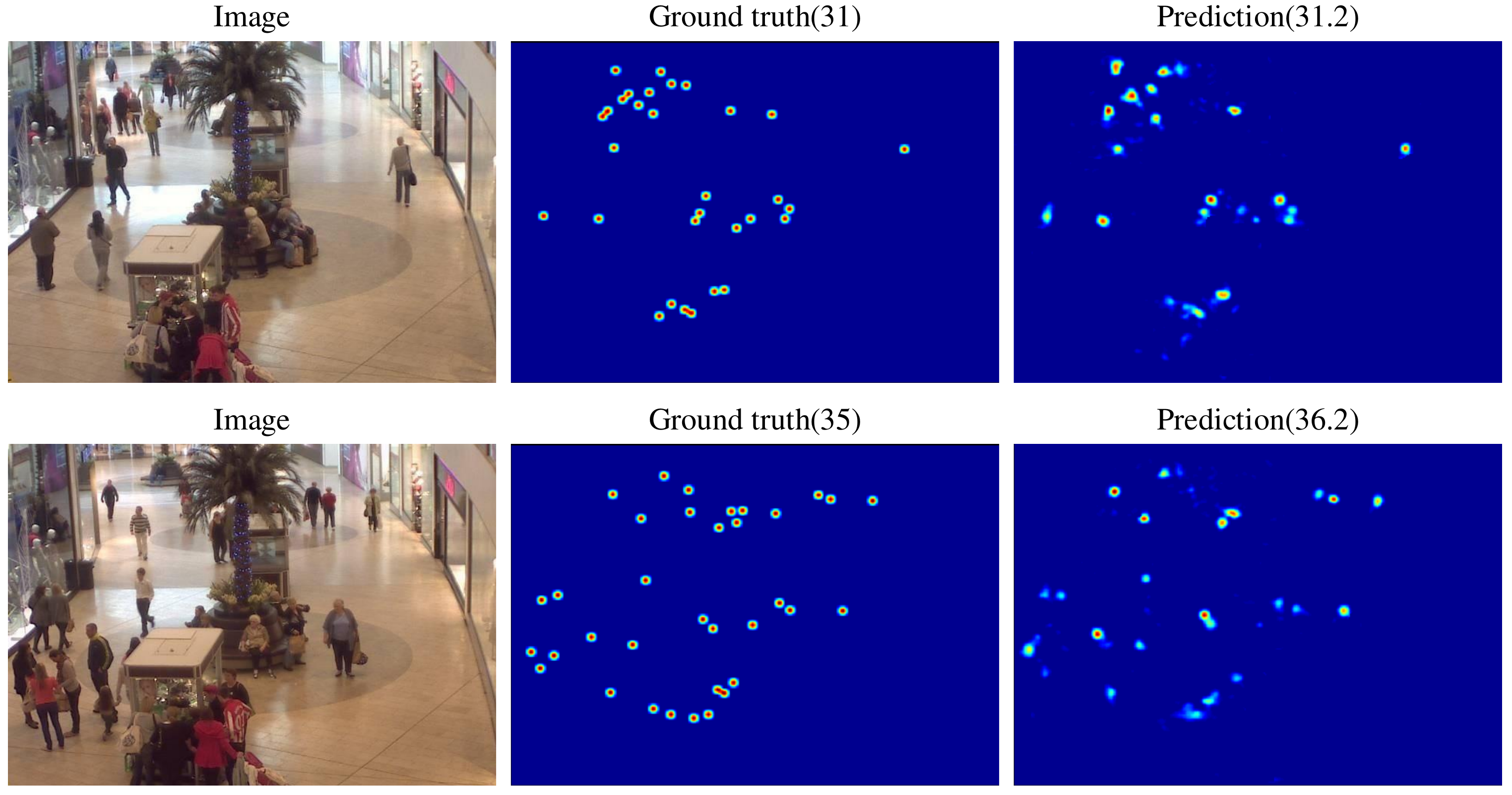}
\caption{Example results on Mall dataset using our ACM-CNN.}
\label{mall_density}
\end{figure}

\subsection{Mall dataset}
The Mall dataset \cite{chen2012feature} is captured in a shopping
mall using a publicly accessible surveillance camera, which suffers
from more challenging lighting conditions and glass surface
reflections. It consists of scenes with more diverse crowd densities
from sparse to crowd, as well as different activity patterns under
different illumination conditions during the day. The video sequence
in the dataset is made up of 2000 frames of resolution 320*240 with
6000 instances of labeled pedestrians. Following the existing
methods, we use the first 800 frames for training and the remaining
1200 frames for evaluation. During testing, MAE is only computed for
regions inside ROI. We perform comparison against Gaussian process
regression \cite{chan2008privacy}, Ridge regression
\cite{chen2012feature}, MoCNN \cite{DBLP:journals/corr/KumagaiHK17},
Count forest \cite{Pham2015COUNT}, and achieves state-of-the-art
performance shown in Table \ref{mall}. Examples are presented in
Figure \ref{mall_density}.

\subsection{UCF\_CC\_50 dataset}
\begin{figure}
\centering
\includegraphics[width=8.8cm]{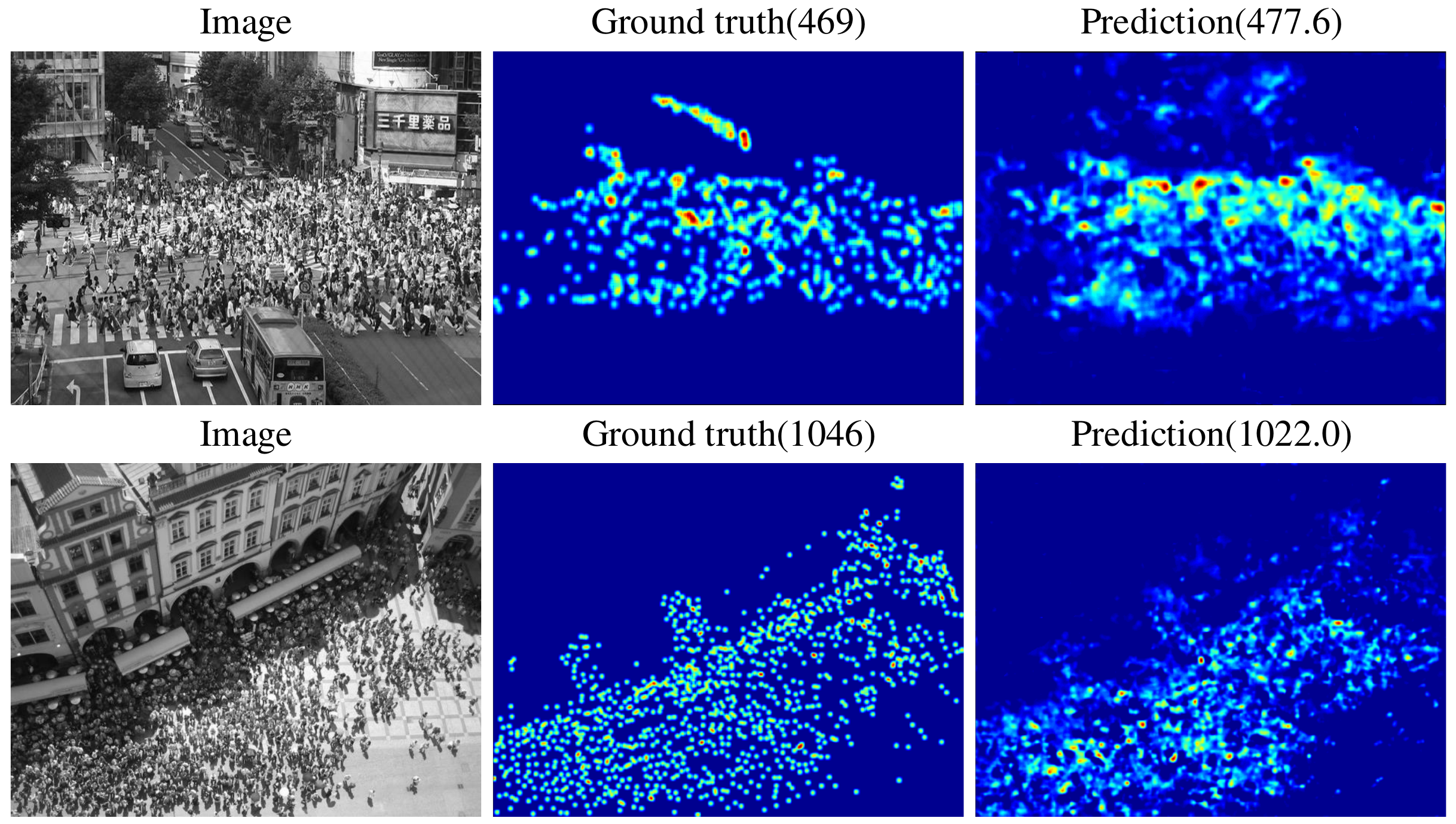}
\caption{Performance of different methods on UCF\_CC\_50 dataset.}
\label{ucf_density}
\end{figure}

The UCF\_CC\_50 dataset \cite{Idrees_2013_CVPR} includes 50 images
with a wide range of densities and diverse scenes. This dataset is
very challenging for the small size and the large variation in crowd
count. As there is no separate test set, a 5-fold cross-validation
method is defined for training and testing to verify the performance
of the model. We compare the proposed model with Multi-source
multi-scale \cite{Idrees_2013_CVPR}, MCNN \cite{zhang2016single},
Switch-CNN \cite{sam2017switching}, CP-CNN \cite{Sindagi_2017_ICCV},
SaCNN \cite{DBLP:journals/corr/abs-1711-04433} and our model
achieves the best MAE over existing methods, shown in Table
\ref{UCF}. Figure \ref{ucf_density} shows the generated density maps
by our ACM-CNN.

\begin{table}[!htb]
\newcommand{\tabincell}[2]{\begin{tabular}{@{}#1@{}}#2\end{tabular}}
% \footnotesize
%\renewcommand{\arraystretch}{1.3}
\caption{Performance evaluation of various methods on UCF\_CC\_50
dataset \cite{Idrees_2013_CVPR}.}\label{UCF} \centering
\begin{tabular}{c|c|c}
\hline
Method & MAE & MSE \\
\hline \hline
Multi-source multi-scale & 468.0 & 590.3 \\
\hline
MCNN & 377.6 & 509.1\\
\hline
Switch-CNN & 318.1 & 439.2 \\
\hline
CP-CNN & 295.8 & \textbf{320.9} \\
\hline
SaCNN & 314.9 & 424.8 \\
\hline
ACM-CNN (ours) & \textbf{291.6} & 337 \\
\hline

\end{tabular}
\end{table}

\section{Analysis}

\begin{figure*}[!htb]
\centering
 \subfigure[Shanghaitech Part\_A]{
 \label{a}
\includegraphics[width=7cm,height=5cm]{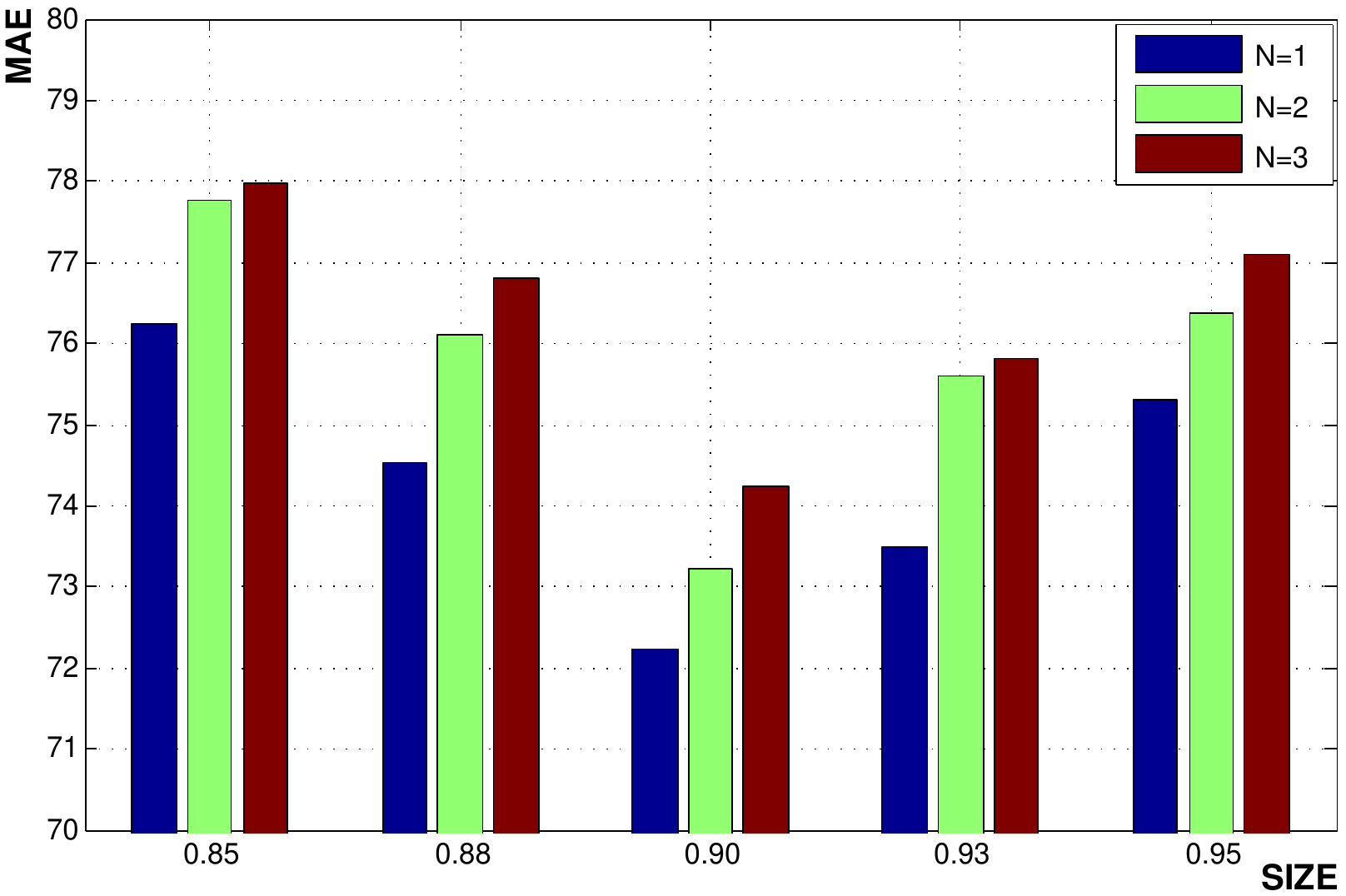}}
\hspace{0.01in}
 \subfigure[UCSD]{
 \label{b}
\includegraphics[width=7cm,height=5cm]{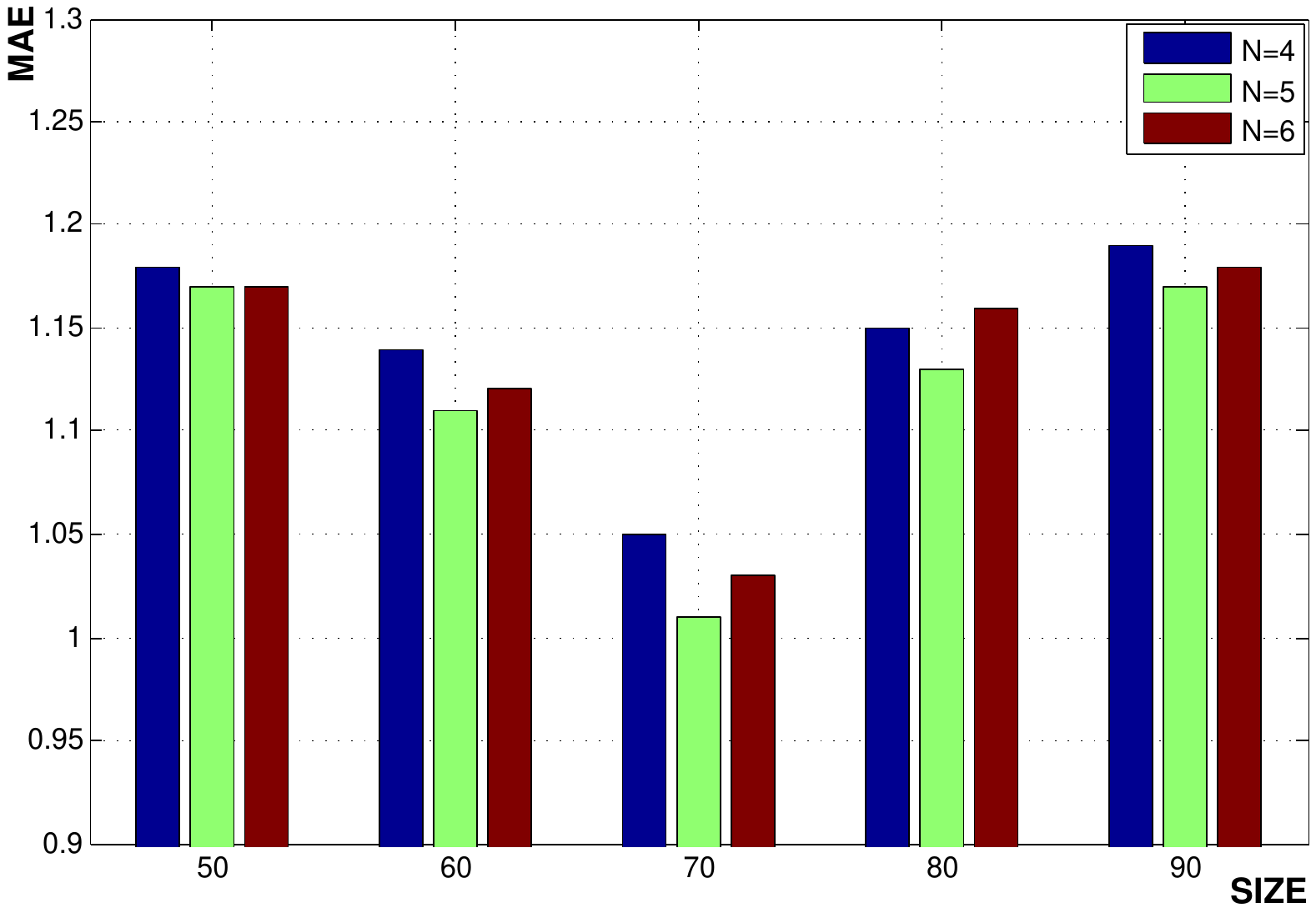}}
\hspace{0.01in} \caption{Histograms of two datasets on parameter
selection. The entire chart is built according to the changes in the
two parameters: the number of attention pixels N and the size of the
corresponding patch S. The left is the Shanghaitech Part\_A dataset
while the right is the UCSD dataset.} \label{parameter}
\vspace{-.3em}
\end{figure*}

\subsection{Parameter Settings}

We first empirically study the configuration of the number of
attention pixels N and the size of the corresponding patch S, as is
shown in Figure \ref{parameter}. Due to the huge diversity in
density levels, there are no set of parameters to meet the
requirements of balancing the performance of all datasets. Thus, we
divide the situation into two types: dense datasets in which all
areas of each image are densely populated and sparse datasets where
only a small portion of each image are occupied by crowds. As is
shown in Table \ref{datasets}, we set a threshold T = 40. When the
average crowd count of one dataset exceeds T, it will be classified
as a dense dataset, otherwise a sparse dataset. Comparative
experiments are respectively performed on two types of
representative datasets: Shanghaitech Part\_A dataset (dense) and
UCSD dataset (sparse). In Figure \ref{a}, the size represents the
ratio of the patch size to the original image's height and width,
whereas in Figure \ref{b}, the size indicates the true pixel value.
This is because images in UCSD dataset share the same size but
Shanghaitech dataset doesn't conform to the condition. According to
the final results, we set N=1 and S= [0.9*height,0.9*width] for
Shanghai Part\_A dataset, N=5 and S = [70,70]
([0.4*height,0.3*width]) for UCSD datasets. The parameters settings
of five datasets used in this paper are listed in Table
\ref{configure}.

\begin{table}[!htbp]
\newcommand{\tabincell}[2]{\begin{tabular}{@{}#1@{}}#2\end{tabular}}
% \footnotesize
%\renewcommand{\arraystretch}{1.3}
\caption{Summary of existing datasets. Max is the maximum crowd
count while Min is minimal crowd count. Ave indicates average crowd
count.}\label{datasets} \centering
\begin{tabular}{c|c|c|c|c}
\hline
\multicolumn{2}{c|}{Dataset} & Max & Min & Ave \\
\hline \hline
\multirow{2}{*}{Shanghaitech} & Part\_A & 3139 & 33 & 501.4\\
\cline{2-5}
& Part\_B & 578 & 9 & 123.6 \\
\hline
\multicolumn{2}{c|}{WorldExpo'10} & 253 &1 & 50.2 \\
\hline
\multicolumn{2}{c|}{UCSD} & 46 & 11 & 24.9 \\
\hline
\multicolumn{2}{c|}{Mall} & 53 & 13 & 31.7 \\
\hline
\multicolumn{2}{c|}{UCF\_CC\_50} & 4543 & 94 & 1279.5 \\
\hline
\end{tabular}
\end{table}

\begin{table}[!htbp]
\newcommand{\tabincell}[2]{\begin{tabular}{@{}#1@{}}#2\end{tabular}}
% \footnotesize
%\renewcommand{\arraystretch}{1.3}
\caption{The parameter settings of five datasets about the number of
attention pixels N and the size of corresponding patch
S.}\label{configure} \centering
\begin{tabular}{c|c|c}
\hline
Type & Datasets & Configure \\
\hline \hline
Dense & \tabincell{c}{Shanghaitech \\ WorldExpo'10 \\ UCF\_CC\_50} & N=1 S=[0.9*height,0.9*width] \\
\hline
Sparse & \tabincell{c}{UCSD \\ Mall} & N=5 S=[0.4*height,0.3*width] \\
\hline
\end{tabular}
\end{table}

\subsection{Algorithmic Study}

\begin{table}[!htb]
\newcommand{\tabincell}[2]{\begin{tabular}{@{}#1@{}}#2\end{tabular}}
% \footnotesize
%\renewcommand{\arraystretch}{1.3}
\caption{Effectiveness of different modules and count attention
mechanism on Part\_A of Shanghaitech and UCSD dataset. C, F, S
correspond to coarse network, fine network and smooth network
respectively.}\label{algorithmic} \centering
\begin{tabular}{c|c|c|c|c}
\hline
\multirow{2}{*}{Method}&\multicolumn{2}{c|}{Shanghaitech}&\multicolumn{2}{c}{UCSD}\\
\cline{2-5}
&MAE&MSE&MAE&MSE\\
\hline \hline
C &106.2&168.4&1.60&2.30\\
\hline
F &80.9&128.1&2.30&3.50\\
\hline
C+F &74.5&108.2&1.19&1.60\\
\hline
C+F+S&\textbf{72.2}&\textbf{103.5}&\textbf{1.01}&\textbf{1.29}\\
\hline
\end{tabular}
\end{table}

\begin{figure*}
\centering
\includegraphics[width=18cm]{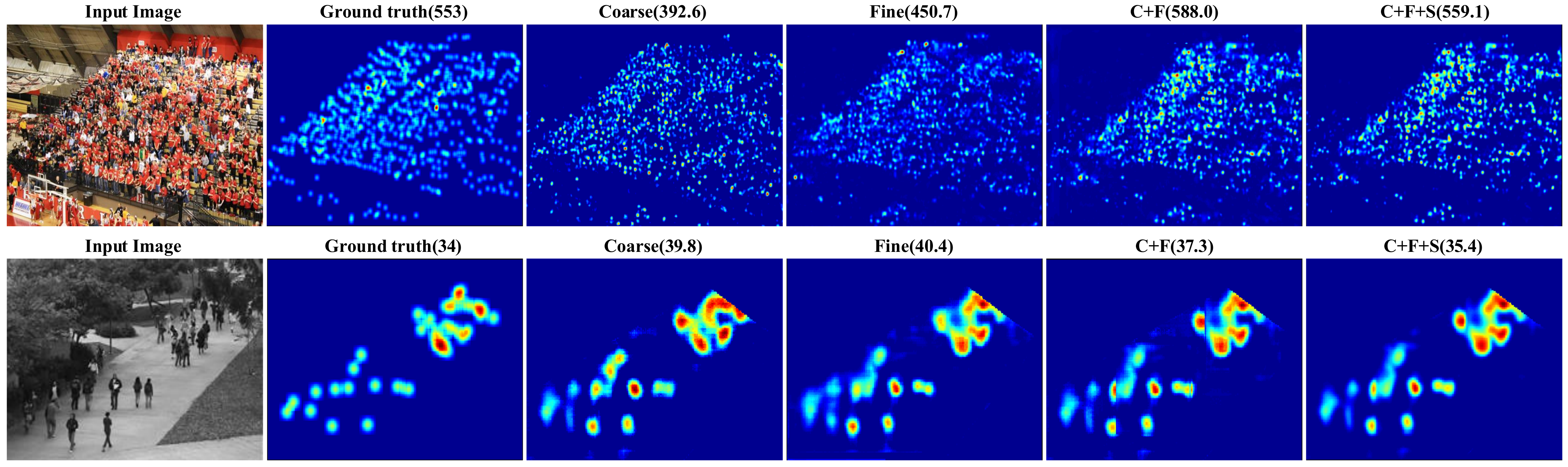}
\caption{Predictions {\color{red}{of}} various structures introduced
in Table \ref{algorithmic} on Shanghaitech Part\_A and UCSD
dataset.} \label{comparative}
\end{figure*}

In this section, we study the effectiveness of modules in the
proposed ACM-CNN and count attention mechanism on the final
accuracy. All ablations are performed on Part\_A of Shanghaitech and
UCSD dataset as they represent two types of datasets: dense and
sparse. First, the first two rows in Table \ref{algorithmic} list
the performance of only coarse network (denoted as C) or only fine
network (denoted as F) on two datasets, which further validate the
previous standpoint that a deep network is suitable for dense scenes
while a shallow network performs better in sparse scenes. In order
to demonstrate the effect of count attention mechanism, we combine
the coarse network and the fine network via this strategy (denoted
as C+F). It is obvious that there is a significant improvement on
two datasets for such a combination compared to using any one of the
modules separately. When introducing the smooth network (defined as
C+F+S), the estimation error is further reduced. This means that the
smooth network is not only able to alleviate the sense of
fragmentation in the result, but also capable of improving the
model accuracy. Figure \ref{comparative} shows the density maps
predicted by various networks in Table \ref{algorithmic} along with
their corresponding ground truths. We can see that the coarse
network yields density maps with fine-grained distribution and the
fine networks are more eager to achieve overall accuracy. Besides,
there is indeed a clear division in the results generated by the
integration of two networks, and the introduction of the smooth
network resolves the problem to some extent.

\subsection{Model generality}
The proposed model is a general framework which means that the
effectiveness of the algorithm does not depend on the choice of
sub-networks. To verify this generality, we train a variant of our
model using the modules introduced in CrowdNet
\cite{Boominathan:2016:CDC:2964284.2967300}. Specifically, the deep
network and shallow network in CrowdNet serve as the fine network
and the coarse network in our model respectively. Without the smooth
network, we perform the experiments on the UCSD datasets and the
results are listed in Table \ref{crowdnet}.

\begin{table}[!htb]
\newcommand{\tabincell}[2]{\begin{tabular}{@{}#1@{}}#2\end{tabular}}
% \footnotesize
%\renewcommand{\arraystretch}{1.3}
\caption{Generality of count attention mechanism using sub-networks
in CrowdNet \cite{Boominathan:2016:CDC:2964284.2967300} on UCSD
dataset.}\label{crowdnet} \centering
\begin{tabular}{c|c|c}
\hline
Method & MAE & MSE \\
\hline \hline
Shallow Network & 1.7 & 2.1 \\
\hline
Deep Network&2.2&2.9\\
\hline
Deep + Shallow (CrowdNet contact) &1.5&1.7 \\
\hline
Deep + Shallow (Count attention) &\textbf{1.2}&\textbf{1.4}\\
\hline

\end{tabular}
\end{table}

CrowdNet concatenates the predictions from the deep and shallow
networks to produce the final density map. From the table, we can
see that it is more effective than using any of these networks
alone. However, combining the two networks with our attention
mechanism is able to achieve better performance, which expresses the
generality of our model. In conclusion, our whole structure is not
limited to the choice of sub-networks. It is a general crowd
counting system to greatly improve the accuracy of predictions.

\section{Conclusion}
In this paper, we propose the Adaptive Multi-scale convolutional
networks, which can assign different capacities to different
portions of the input and characterize them with corresponding
networks. Our model consists of three modules, including the coarse
network, the fine network and the smooth network. It can be regarded
as a general crowd counting system since the choice of three types
of sub-networks is flexible. Extensive experiments on five
representative datasets (three dense datasets and two sparse
datasets) demonstrate that the proposed model delivers
state-of-the-art performance over existing methods. Further, we
validate the effectiveness of modules and the attention mechanism
with ablations. In the future, we would like to consider extending
our method to fit for other scenario
\cite{Cheng:2014:TSM,Wang_2016_CVPR}.

\section*{References}

\bibliography{mybibfile}

\end{document}